\RequirePackage{fix-cm}
\documentclass[twocolumn]{svjour3}          
\usepackage{moreverb,url}
\smartqed  
\usepackage{graphicx}
\usepackage{amsmath}
\usepackage{blindtext}
\usepackage{amsfonts}
\usepackage{amssymb}
\usepackage{algorithm,algpseudocode}
\usepackage{lscape}
\usepackage{rotating}

\newcommand{\vp}{\mathbf{p}}
\newcommand{\vt}{\mathbf{t}}
\newcommand{\vu}{\mathbf{u}}
\newcommand{\vv}{\mathbf{v}}

\begin{document}

\title{Solving Viewing Graph Optimization for Simultaneous Position and Rotation Registration 
}



\author{Seyed-Mahdi Nasiri \and
        Reshad Hosseini\and
        Hadi Moradi 
}


\institute{SM. Nasiri \at
              \email{s.m.nasiri@gmail.com}\\           
              \emph{School of ECE, College of Engineering, University of Tehran, Tehran, Iran} \\
           \and
           R. Hosseini \at
              \email{reshad.hosseini@ut.ac.ir}\\
              Tel.: +98-21-82089799\\
              \emph{School of ECE, College of Engineering, University of Tehran, Tehran, Iran}\\
              \emph{School of Computer Science, Institute of Research in Fundamental Sciences (IPM), Tehran, Iran}\\
           \and
           H. Moradi \at
              \email{hadi.moradi@ut.ac.ir}\\
              Tel.: +98-21-82084960\\
              \emph{School of ECE, College of Engineering, University of Tehran, Tehran, Iran}\\
              \emph{Intelligent Systems Research Institute, SKKU, South Korea}
}
\date{Received: date / Accepted: date}

\maketitle

\begin{abstract}
A viewing graph is a set of unknown camera poses, as the vertices, and the observed relative motions, as the edges. Solving the viewing graph is an essential step in a Structure-from-Motion procedure, where a set of relative motions is obtained from a collection of 2D images. Almost all methods in the literature solve for the rotations separately, through rotation averaging process, and use them for solving the positions. Obtaining positions is the challenging part because the translation observations only tell the direction of the motions.
It becomes more challenging when the set of edges comprises pairwise translation observations between either near and far cameras.
In this paper an iterative method is proposed that overcomes these issues. Also a method is proposed which obtains the rotations and positions simultaneously. Experimental results show the-state-of-the-art performance of the proposed methods.
\keywords{Viewing Graph \and Structure-from-Motion \and Camera Pose Registration}
\end{abstract}

\section{Introduction}\label{intro}


Structure-from-Motion (SfM) refers to a process in which a set of 3D points are reconstructed from their projections on a given set of images. Almost all SfM techniques can be classified into two categories, sequential or incremental techniques \cite{snavely2006photo,agarwal2011building,frahm2010building,wu2013towards,furukawa2010towards,havlena2009randomized,snavely2008skeletal}, and global or batch techniques \cite{jiang2013global,moulon2013global,arie2012global,crandall2011discrete,sweeney2015optimizing}.

As the incremental approaches often suffer from large drifting error, it is recommended to use global methods to achieve better accuracy and consistent models.
The global methods consist of the following main steps:
1) Pairwise image registration: Estimating relative rotations and movement directions between pairs of images using feature points \cite{torr2000mlesac,nister2004efficient,kukelova2008polynomial}. The result is a graph, i.e. the viewing graph, whose edges are the relative pose measurements. 2) Solving the viewing graph: Camera poses estimation \cite{ozyesil2015robust,sweeney2015optimizing,jiang2013global,zhu2018very,hartley2013rotation,chatterjee2017robust,arrigoni2018robust}, which is the topic of this paper. 3) Triangulation: Reconstructing 3D points by triangulating corresponding points from calculated camera points \cite{hartley1997triangulation,stewenius2005hard,byrod2007improving}. 4) Bundle adjustment: Camera poses and 3D points refinement by reprojection error minimization \cite{triggs1999bundle,lourakis2009sba}.

Since the relative translation observations only contains the movement directions and not their scale, solving the viewing graph is the challenging step in the SfM process. To simplify the problem, almost all viewing graph solvers determine the rotations using only the rotation observations and then use them to compute the camera positions \cite{wilson2014robust,govindu2001combining,ozyesil2015robust,jiang2013global}. Obtaining a set of rotations from their relative observations is known as rotation averaging problem and has been studied in the computer vision community \cite{nasiri2018linear,hartley2011l1,arrigoni2014robust,hartley2013rotation}.

The rotations obtained by solving the rotation averaging problem are used to change the representation of direction observations from local coordinates to a global coordinate system. The camera location estimation problem is to find camera positions from these relative directions in the global coordinate frame. A solution is to find camera locations which minimize the squared sum of direction errors \cite{wilson2014robust}. 
The difficulty of solving this non-linear cost function led most of the available methods in the literature to change the cost function from the direction error to the displacement error.
To this end, unknown scale factors are added to the problem which should be estimated besides the camera poses.
Some of these approaches used the cross product of solution translations with direction observations as displacement error \cite{govindu2001combining,arie2012global}.
These approaches suffer from the property of cross product distance that decreases for an angle error of more than $90$ degrees.
The constrained least-squared \cite{tron2014distributed,tron2009distributed}, and least-unsquared chordal distances \cite{ozyesil2015stable,ozyesil2015robust} are other commonly used cost functions. Using the displacement error instead of direction error biases the errors in proportion to the length of edges.

In this paper, we show that the constrained least-squared formulation for solving camera positions has inherit limitation that decreases its performance.  We, therefore, propose an iterative algorithm to solve the original displacement error or direction error cost functions starting from the solution of the constrained least-squared formulation. Experimental results show the efficacy of the proposed methods. We also propose iterative methods for solving simultaneous position and rotation registration. These methods solves a pose graph optimization (PGO) problem in each step, that thanks to recent success of PGO solvers can be solved efficiently and with high accuracy. Experimental results show that our proposed methods significantly outperform most accurate methods proposed in the literature.



\section{Preliminaries}\label{sec:Pre}
\subsection{Viewing-Graph}

Given $n$ views or images from a scene, a subset of $\binom{n}{2}$ pairwise relative motions can be estimated.
The camera poses, as the vertices, and the pairwise relative motions, as the observed edges, construct the viewing graph. Each relative motion observation comprises a relative rotation and a relative direction. Relative directions are vectors from an endpoint of an edge to another one and are represented in local coordinate frames of vertices.

A set of vertices, $V$, and its relevant set of edges, $E$, form a viewing graph $G=\{V,E\}$. The parameters of the $i^{th}$ vertex, $v_i$, are the position and the orientation of the $i^{th}$ camera with respect to a global coordinate frame, i.e. $v_i=\{\vp_i,R_i\}$.  An edge between the $i^{th}$ and $j^{th}$ vertices, represented by $e_{ij}$, comprise a noisy measurement of the relative rotation, $R_{ij} \leftarrow R_j^T R_i$, and a noisy measurement of the direction of movement in the $i^{th}$ coordinate frame, i.e. $\gamma^l_{ij} \leftarrow R_i^T(\vp_j-\vp_i)/\|\vp_j-\vp_i\|$.

Solving a viewing graph means finding unknown parameters of the vertices, i.e. camera poses, that match the edges, i.e. relative observations, as much as possible. Mathematically speaking, the viewing graph optimization problem solves for the edges that minimize the following mismatch cost function,
\begin{equation}\label{cost}
    \min_{\{\vp_i,R_i\}} \sum_{e_{ij}=\{i,j\}\in E} d^2_R\left(R_{ij},R_j^T R_i\right) + d^2_{d}(\gamma_{ij},\frac{\vp_j-\vp_i}{\|\vp_j-\vp_i\|}),
\end{equation}
where, $d_R(.)$ and $d_\gamma(.)$ are distance metrics and $\gamma_{ij} = R_i\gamma^l_{ij}$. The weights can be added to the cost function to handle different confidence level of measurements and to balance errors of the two parts of the cost function.

\subsection{Common Metrics}

Rotations distance metrics are widely studied under the title of \textit{rotation averaging} \cite{hartley2013rotation,chatterjee2017robust}.
The common metric, that is also used in our proposed methods, is the Frobenius norm of difference of the rotation matrices:
\begin{equation}\label{rot_cost}
    d_R\left(R_1,R_2\right) = \left\|R_1-R_2\right\|_F.
\end{equation}

Several distance metrics can also be used for the direction part of the cost function, i.e. $d_d(.)$. Some cases are listed in \cite{wilson2014robust}. The commonly used metrics are the chordal and the orthogonal distances which are formulated as follows:
\begin{align}
\label{chord}
    \text{Chordal: }& d_{d}\left(\vu,\vv\right)=\left\|\vu-\vv\right\|\\
\label{crs}
    \text{Orthogonal: }& d_{d}\left(\vu,\vv\right)=\left\|{P}_{\vu^\perp}(\vv)\right\|=\left\|\vu\times\vv\right\|
\end{align}
where ${P}_{\vu^\perp}$ is the projector onto the orthogonal complement of the span of $\vu$ and ``$\times$'' represents the cross product.

Common approaches separate the two parts of the cost function. The first part, i.e. the rotation averaging problem \cite{hartley2013rotation,nasiri2018linear}, is independent of the positions and can be solved independently to obtain the vertices orientations.
The rotation averaging problem is defined as:
\begin{equation}\label{rot_avg}
    \min_{\{R_i\}} \sum_{e_{ij}=\{i,j\}\in E} d^2_R\left(R_{ij},R_j^T R_i\right).
\end{equation}

The obtained orientations are used to calculate $\gamma_{ij}$s from $\gamma^l_{ij}$s. Now the second part of the cost function is a function of positions and known as ``camera location estimation'' problem. The objective function is designed to minimize directions error and is formulated as:
\begin{equation}\label{CLE1}
    \min_{\{\vp_i\}} \sum_{e_{ij}=\{i,j\}\in E} d^2_{d}(\gamma_{ij},\frac{\vp_j-\vp_i}{\|\vp_j-\vp_i\|}).
\end{equation}

There are global scale and translation ambiguities in the solutions of \eqref{CLE1}, and if optimization methods can cope with over-parametrization there is no need to remove the ambiguities. Some existing methods however use some constraints to resolve the ambiguites. The constraints $\sum \vp_i =0$ or fixing the first vertex at the origin are the common constraints to remove the translation ambiguity, and the constraints like $\sum \|\vp_i\|^2=1$ or $\|\vp_i-\vp_j\|>1,\forall (i,j)\in E$ can be used to remove the scale ambiguity.

A different approach to find camera locations is to minimize the displacements error instead of directions error. In this case the problem is formulated as:
\begin{equation}\label{CLET}
    \min_{\{\vp_i\}} \sum_{e_{ij}=\{i,j\}\in E} d^2_{d}(\|\vp_j-\vp_i\|\gamma_{ij},\vp_j-\vp_i).
\end{equation}

In this case. it is important to consider a constraint like $\sum \|\vp_i\|^2=1$ to prevent the trivial solutions of placing all camera positions at the same point.

\begin{figure}
    \includegraphics[width=0.5\textwidth]{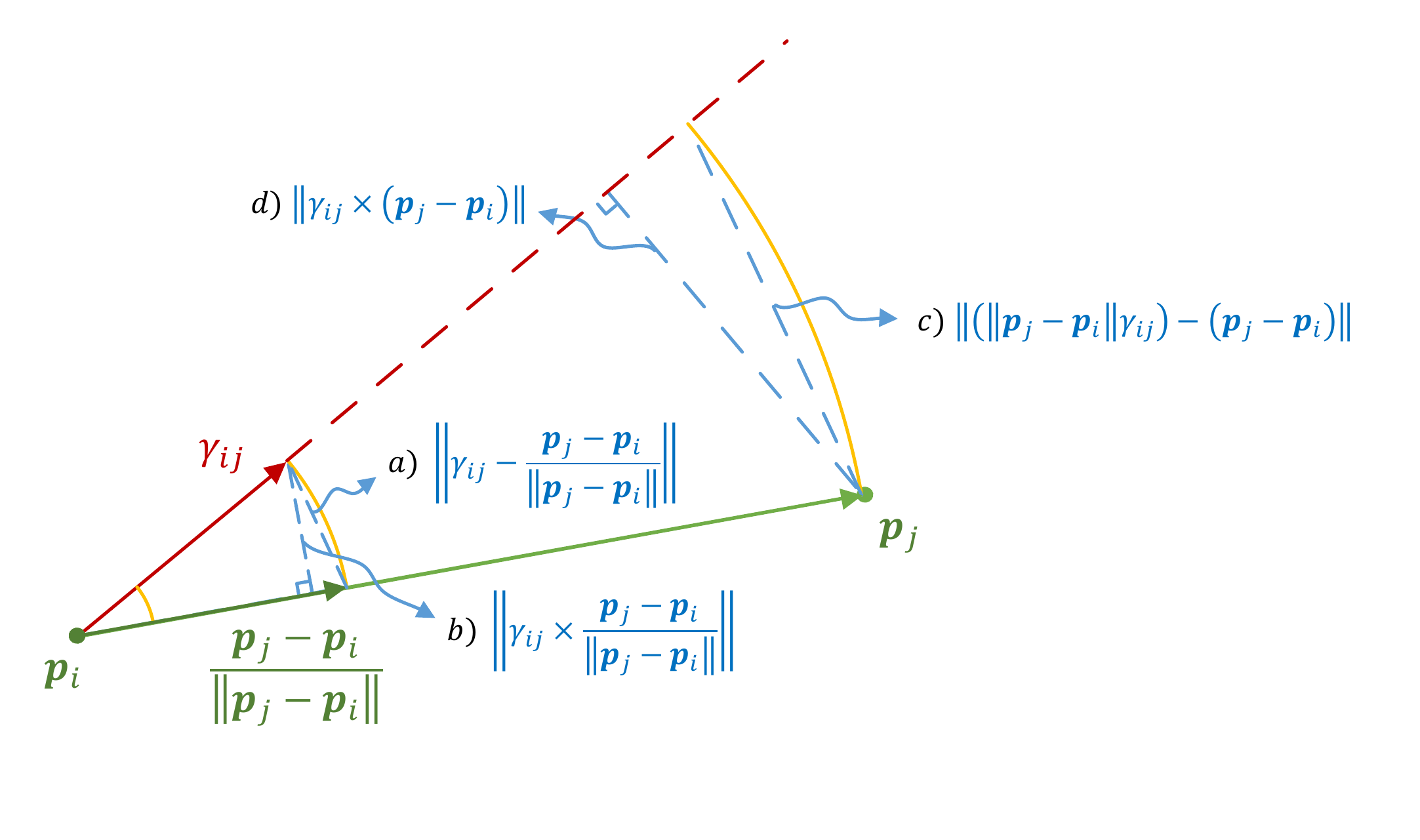}
\caption{The geometric interpretation of different error criteria. a) Chordal distance of directions error. b) Orthogonal distance of directions error. c) Chordal distance of displacements error. d) Orthogonal distance of displacements error.}
\label{fig:metrics}       
\end{figure}

The combinations of two distance metrics, i.e. orthogonal and chordal, and the two aforementioned approaches, i.e. minimizing directions or displacements error, form a set of four different error criteria which were used by different methods to solve the camera location estimation problem.
The geometric interpretation of these error criteria is shown in Fig.~\ref{fig:metrics}.

\section{Limitations of Existing Methods}
In this section, we first talk about a limitation of an existing method to solve the cost function of~\eqref{CLET}. The method is fast and gives relatively good results and we use it as the initialization procedure of our proposed methods. In the next part, we show the limitation of the orthogonal distance, and therefore the existing methods in the literature that use this cost function.

\subsection{Linear constraint problem}\label{conswaek}
A common approach of solving camera locations is to solve the following alternative problem:
\begin{equation}\label{CLE2}
    \min_{\{\vp_i,\lambda_{ij}\}} \sum_{e_{ij}=\{i,j\}\in E} \left\|\lambda_{ij}\gamma_{ij}-(\vp_j-\vp_i)\right\|^2,
\end{equation}
where the scale factors, $\lambda_{ij}$s, are unknown parameters. For $\lambda_{ij} = \|\vp_j-\vp_i\|$, problem \eqref{CLE2} is the same as \eqref{CLET} with the chordal distance, which minimizes the displacement error, i.e. Euclidean distance between the endpoint of vectors $\lambda_{ij}\gamma_{ij}$ and $\vp_j-\vp_i$.
In practice, the constraint $\lambda_{ij} = \|\vp_j-\vp_i\|$ is dropped and the linear constraint $\lambda_{ij}\geq 1$ is replaced, which also handles the scale ambiguity. Like \cite{arrigoni2014robust}, we refer to the problem of \eqref{CLE2} with the constraint $\lambda_{ij}\geq 1$ as the \textit{constrained least-squares (CLS)}.

The cost function of \eqref{CLE2} is quadratically related to the global scale, i.e. the size of the vertex set. Reducing the global scale reduces the cost quadratically. Obviously, setting the global scale to zero, results in zero cost with the trivial solution of $\{\vp_i=0, \forall i\in V\}$.
However, the constraint $\lambda_{ij}\geq 1$, prevents a solver to get this trivial solution.
In fact, $\lambda_{ij}$s represent the distance between the vertices, and the constraint prevents the vertices from getting too close to each other.

At a first glance, it seems that the constraint causes the global scale to be determined according to the minimum distance between the vertices. This implies that the smallest scale factor, i.e. the minimum $\lambda_{ij}$ corresponding to the minimum distance of the vertices, is set to one and the other scale factors are estimations of the distance of their endpoints divided by the minimum distance of the vertices.
But in practice, a significant number of $\lambda_{ij}$s are exactly set to one. In fact, $\lambda_{ij}$s for a subset of edges, with shorter length than the others, are set to one. This will increase the cost on the subset of edges, but will reduce the global scale and reduce the cost on the other edges with longer length. Therefore, the cost is decreased, but $\lambda_{ij}$s and consequently the positions deviate from their real values.

\begin{figure}
    \includegraphics[width=0.5\textwidth]{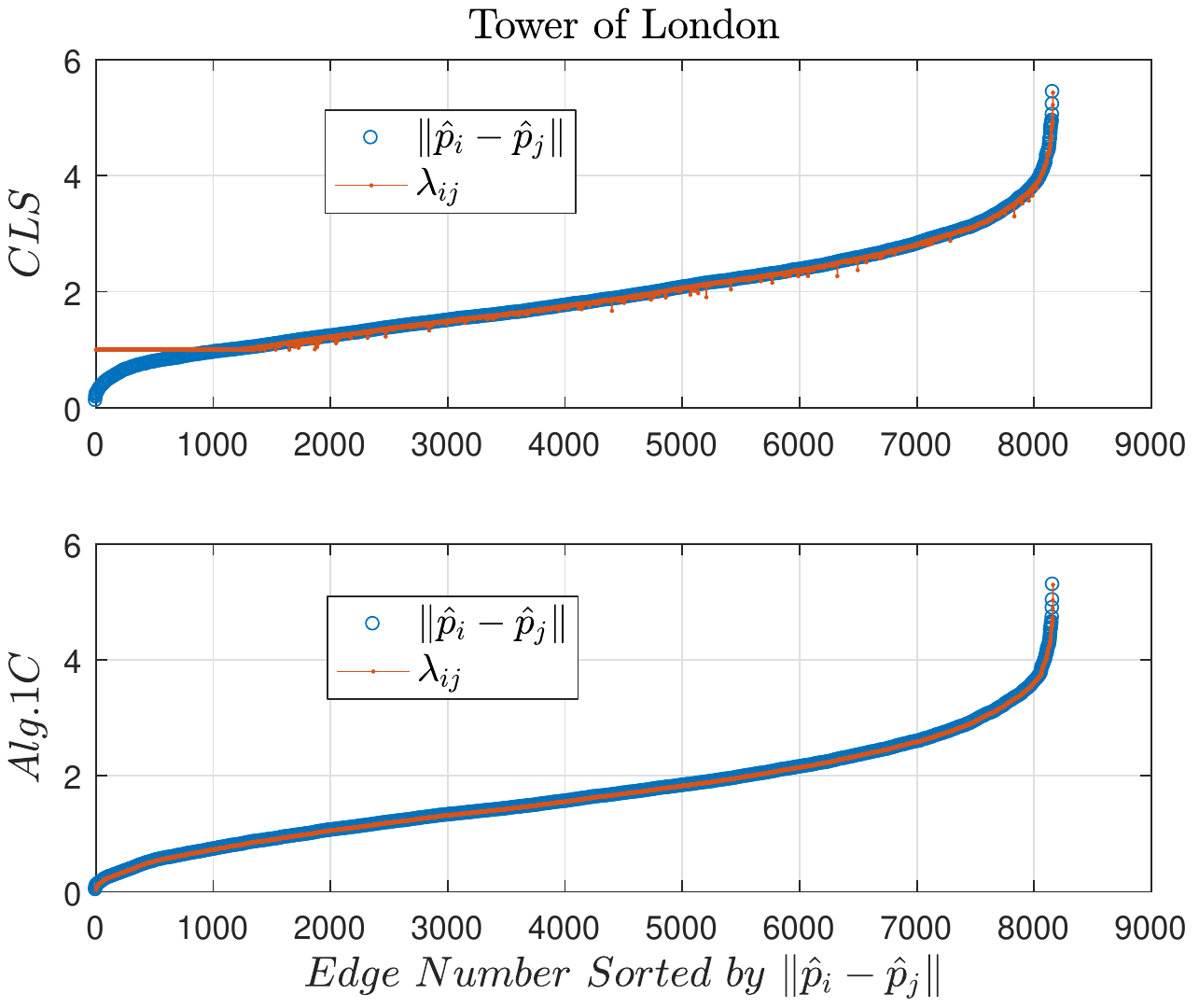}
\caption{Comparing the obtained scale factors, i.e. $\lambda_{ij}$s, in CLS and Alg.~1C on ``Tower of London'' dataset. A significant number of scale factors, obtained by CLS, are stick to the lower bound and set to $1$. That is while the scale factors obtained by Alg.~\ref{algorithm1}C match the distances between estimated camera positions.}
\label{fig:CLS_Alg1C_wD2}       
\end{figure}

Fig.~\ref{fig:CLS_Alg1C_wD2} compares the obtained $\lambda_{ij}$s by CLS and one of our proposed methods for a dataset. 
In this figure, there are a large number of vertices with small $\|\hat{\vp}_j-\hat{\vp}_i\|$, such that $\lambda_{ij}$ is set to $1$ and therefore does not match $\|\hat{\vp}_j-\hat{\vp}_i\|$.
Furthermore, it shows that for CLS the mismatches also occurred in large $\lambda_{ij}$s. In contrast in our proposed method, $\lambda_{ij}$s and their corresponding $\|\hat{\vp}_j-\hat{\vp}_i\|$, completely match.

\subsection{Orthogonal metric problem}
Orthogonal distance is a common distance metric used in the literature for solving the camera locations. It is actually an easier metric to be solved and one can have interesting convergence results. It can be shown that the following cost function that has multipliers $\lambda_{ij}$ can be solved iteratively and at the minimum is equal to the orthogonal distance metric (see Appendix for more details).
\begin{equation}\label{Crs3}
    \min_{\{\vp_i,\lambda_{ij}\}} \sum_{e_{ij}=\{i,j\}\in E} \left\|\gamma_{ij}-\lambda_{ij}(\vp_j-\vp_i)\right\|^2.
\end{equation}

The orthogonal distance decreases for an angle error of more than 90 degrees.
This means that the angle between $\gamma_{ij}$ and $(\vp_j-\vp_i)$, and also the Euclidean distance between them, can be increased while their orthogonal distance is reduced.
This can results in undesirable results in methods based on this distance metric. 

\begin{figure}
    \label{fig:graph}
    \includegraphics[width=0.23\textwidth]{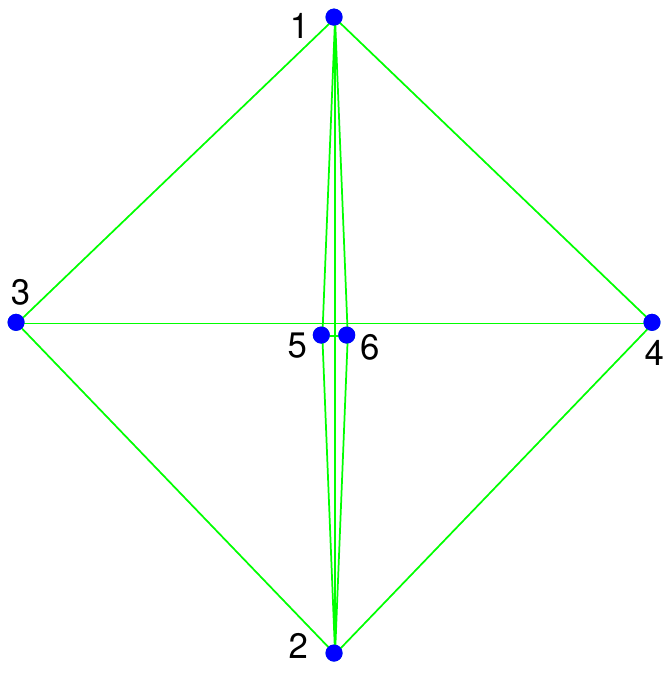}
    \includegraphics[width=0.23\textwidth]{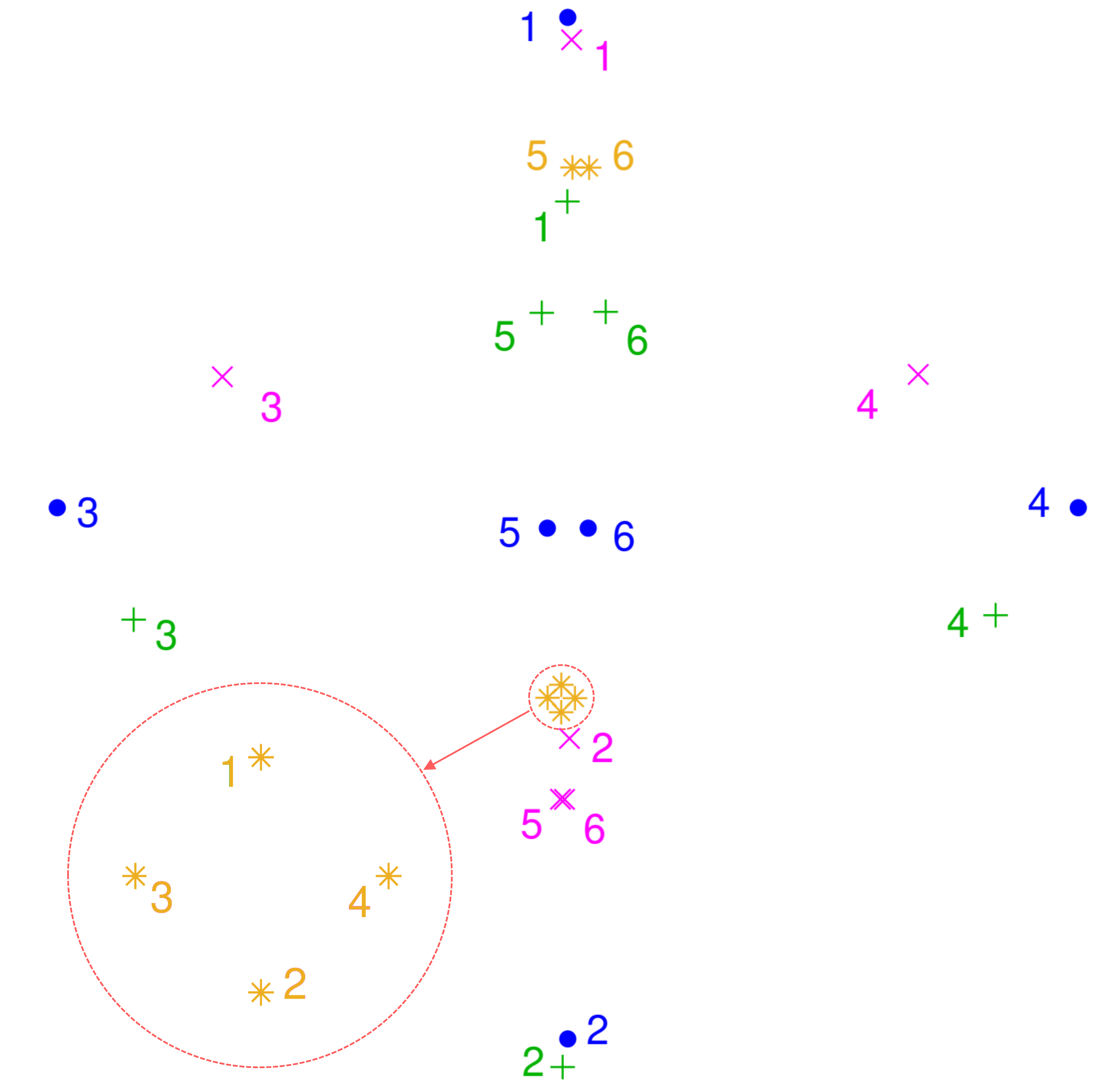}
    \caption{The results of applying algorithms CLS (green plus), the cross product minimization of \cite{govindu2001combining} (purple cross), and orthogonal projection minimization of \cite{goldstein2016shapefit} (orange stars). The graph vertices and edges are shown as blue dots and green lines.}
\end{figure}

We construct a simple example to demonstrate the problem of orthogonal distance. Fig.~\ref{fig:graph} shows a graph with six vertices. Vertices number $5$ and $6$, placed near the center of the graph, are closer to each other than the other vertices. The rotation observations assumed to be exact and the direction observations are perturbed by $5$ degrees error. The cross product method of \cite{govindu2001combining} and ShapeFit method of \cite{goldstein2016shapefit} and the CLS algorithms are applied to the problem and the results are shown in the figure.
The relative positions of the first four vertices in all three methods are close to their true relative positions, and the four vertices are located in the four vertices of a rhombus.
The positions of vertices number $5$ and $6$ estimated by CLS are far from their true relative positions, but these vertices are still inside the rhombus of the first four vertices. That is while the positions of vertices number $5$ and $6$ estimated by the other two methods is outside of the rhombus in a situation that their positions relative to one of the vertices $1$ and $2$ is in the opposite direction of the real relative direction. Note that, when the angle between the relative position and the direction observation is about $180$ degrees, i.e. they are in opposite directions, then the orthogonal distance between them are very small. Therefore, the cost value obtained by the methods based on the orthogonal distance will be minimized, while the relative positions of some vertices are completely in different position with respect to their real relative positions to the positions of other vertices.


\section{The Proposed Methods}\label{sec:Method}
To overcome the problem of linear constraint weakness, discussed in \ref{conswaek}, we propose an iterative solver used in two different approaches, minimizing sum of squared of displacements and directions error, to solve the viewing graph problem.
The main idea is to use $\lambda_{ij}$s obtained by \textit{CLS} at the first iteration and then replace them with fixed coefficients 
according to the obtained positions in the last iteration. We are not able to prove the convergence of the iterative solvers, but we observe experimentally that all of our iterative solvers converge.

To minimize the sum of squared displacements error, the formulation of \eqref{CLE2} is employed and $\lambda_{ij}$s are set to obtained distance of the endpoints of the vertices $i$ and $j$ in the last iteration of the algorithm.
Therefore upon convergence, the algorithm yields $\lambda_{ij} = \|\hat{\vp}_j-\hat{\vp}_i\|$, where $\hat{\vp_i}$s are the estimated camera positions and we are solving the actual optimization problem of~\eqref{CLET}.

\renewcommand{\algorithmicrequire}{\textbf{Input:}}
\renewcommand{\algorithmicensure}{\textbf{Output:}}
\begin{algorithm}[tbp]
\small
\newcommand{\NewComment}[1]{ {\hfill$//$ #1}} 
\algnewcommand{\IIf}[1]{\State\algorithmicif\ #1\ \algorithmicthen}
\algnewcommand{\EndIIf}{\unskip\ \algorithmicend\ \algorithmicif}
  \caption{}
    \begin{algorithmic}[1]
	    \Require{Edges of a viewing graph ($R_{ij},\gamma^l_{ij},\, \forall \{i,j\}\in E$)
        }

        approach:

        ``C'' minimizing the common used objective function which comprises displacements error.

        ``O'' minimizing the original objective function which comprises directions error.

        \Ensure{Estimated poses of vertices  ($\hat{R}_i,\hat{\vp}_i,\, \forall i\in V$)}
        \State $\hat{R}_i \gets RS(R_{ij})$ \Comment{Find initial orientations by RS}
        \State $\gamma_{ij} \gets \hat{R}_i\gamma^l_{ij},\,  \forall e_k=\{i,j\}\in E$ \Comment{Move directions to global coordinate}
        \State $G'=\{E',V'\} \gets 1DSFM(G=\{E,V\})$ \Comment{Remove outliers by 1DSFM}
        \State $\{\hat{\vp}_i\} \gets CLS(\gamma_{ij})$ \Comment{Solve CLS to obtain positions}
		\Repeat
            \If{approach=``C''}
                \State $\lambda_{ij} \gets \|\hat{\vp}_j-\hat{\vp}_j\|,\,  \forall e'_k=\{i,j\}\in E'$
                \State $\{\hat{\vp}_i\} \gets LS1(\lambda_{ij}\gamma_{ij})$ \Comment{Solve \eqref{CLE2} with fixed $\lambda_{ij}$s to update positions}
            \ElsIf{approach=``O''}
                \State $\lambda_{ij} \gets 1/\|\hat{\vp}_j-\hat{\vp}_j\|,\,  \forall e'_k=\{i,j\}\in E'$
                \State $\{\hat{\vp}_i\} \gets LS2(\lambda_{ij},\gamma_{ij})$ \Comment{Solve \eqref{Crs3} with fixed $\lambda_{ij}$s to update positions}
            \EndIf

        \Until{convergence}
        \State \Return {$\hat{R}_i,\hat{\vp}_i,\, \forall i\in V'$}
    \end{algorithmic}
  \label{algorithm1}
\end{algorithm}

\renewcommand{\algorithmicrequire}{\textbf{Input:}}
\renewcommand{\algorithmicensure}{\textbf{Output:}}
\begin{algorithm}[tbp]
\small
\newcommand{\NewComment}[1]{ {\hfill$//$ #1}} 
\algnewcommand{\IIf}[1]{\State\algorithmicif\ #1\ \algorithmicthen}
\algnewcommand{\EndIIf}{\unskip\ \algorithmicend\ \algorithmicif}
  \caption{}
    \begin{algorithmic}[1]
	    \Require{Edges of a viewing graph ($R_{ij},\gamma^l_{ij},\, \forall e_k=\{i,j\}\in E$)
        }

        approach:

        ``C'' minimizing the common used objective function comprises displacements error.

        ``O'' minimizing the original objective function comprises directions error.

        \Ensure{Estimated poses of vertices  ($\hat{R}_i,\hat{\vp}_i,\, \forall i\in V$)}
        \State $\hat{R}_i \gets RS(R_{ij})$ \Comment{Find initial orientations by RS}
        \State $\gamma_{ij} \gets \hat{R}_i\gamma^l_{ij},\,  \forall e_k=\{i,j\}\in E$ \Comment{Move directions to global coordinate}
        \State $G'=\{E',V'\} \gets 1DSFM(G=\{E,V\})$ \Comment{Remove outliers by 1DSFM}
        \State $\{\hat{\vp}_i\} \gets CLS(\gamma_{ij})$ \Comment{Solve CLS to obtain positions}
		\Repeat
            \State $\gamma_{ij} \gets \hat{R}_i\gamma^l_{ij},\,  \forall e_k=\{i,j\}\in E$ \Comment{Move directions to global coordinate according to the updated rotations}
            \If{approach=``C''}
                \State $\lambda_{ij} \gets \|\hat{\vp}_j-\hat{\vp}_j\|,\,  \forall e'_k=\{i,j\}\in E'$
                \State $w_{ij} \gets 1,\,  \forall e'_k=\{i,j\}\in E'$
            \ElsIf{approach=``O''}
                \State $\lambda_{ij} \gets \|\hat{\vp}_j-\hat{\vp}_j\|,\,  \forall e'_k=\{i,j\}\in E'$
                \State $w_{ij} \gets \frac{1}{\|\hat{\vp}_j-\hat{\vp}_j\|^2},\,  \forall e'_k=\{i,j\}\in E'$
            \EndIf
            \State $\{\hat{\vp}_i,\hat{R}_i\} \gets PS(\lambda_{ij}\gamma^l_{ij},R_{ij},w_{ij})$ \Comment{Use PS \cite{nasiri2020novel} to update positions and orientations}
        \Until{convergence}
        \State \Return {$\hat{R}_i,\hat{\vp}_i,\, \forall i\in V'$}
    \end{algorithmic}
  \label{algorithm2}
\end{algorithm}

The proposed algorithm summarized in Alg.~\ref{algorithm1} consists of the following steps:
\begin{enumerate}
  \item Use the RS algorithm \cite{nasiri2020novel} to solve \eqref{rot_avg} and find the orientation of the vertices. (RS uses the Frobenius norm for $d_R(.)$)
  \item Use 1DSFM algorithm \cite{wilson2014robust} to remove the outliers.
  \item Solve \eqref{CLE2} on the common constraint of $\lambda_{ij}\geq1$ to estimate $\hat{\vp}_i$s.
  \item Replace parameters $\lambda_{ij}$ with fixed coefficients $\lambda_{ij} = \|\hat{\vp}_j-\hat{\vp}_i\|$.
  \item Solve \eqref{CLE2} with fixed $\lambda_{ij}$s, which now is a unconstrained linear least-squares problem. (We refer to it as LS1 in Alg.~\ref{algorithm1})
  \item Repeat the last two steps until convergence.
\end{enumerate}




To minimize the directions error instead of displacements error, we can use the formulation of \eqref{Crs3}. Therefore the $4^{th}$ and $5^{th}$ step of the algorithm is replaced by:

\begin{enumerate}
    \setcounter{enumi}{3}
  \item Replace parameters $\lambda_{ij}$ with fixed coefficients $\lambda_{ij} = 1/\|\hat{\vp}_j-\hat{\vp}_i\|$.
  \item Solve \eqref{Crs3} with fixed $\lambda_{ij}$s, which now is a unconstrained linear least-squares problem. (We refer to it as LS2 in Alg.~\ref{algorithm1})
\end{enumerate}


Fixing the parameters $\lambda_{ij}$s converts the direction observations to translation observations. This means that the given viewing graph is converted to a Pose-Graph, and the problem is turned to a Pose-Graph optimization problem:
\begin{equation}\label{PGO}
\begin{split}
    \min_{\{\vp_i,R_i\}} \sum_{e_{ij}=\{i,j\}\in E} &\left\|R_{ij},R_j^T R_i\right\|_F^2 +\\
     &w_{ij}\left\|\lambda_{ij}\gamma_{ij}-(\vp_j-\vp_i)\right\|^2.
     \end{split}
\end{equation}

The literature of the Pose-Graph optimization has a wide variety of algorithms from which there are algorithms that solve the the orientations and positions of the vertices simultaneously \cite{nasiri2020novel,rosen2019se}. This was the motivation of our second method which optimizes the positions and the orientations together. The idea is to change the $5^{th}$ step of the algorithm with the state-of-the-art PS algorithm proposed in \cite{nasiri2020novel} to find the vertices' poses.
The proposed algorithm is summarized in Alg.~\ref{algorithm2}.

In both of our proposed algorithm, we use displacement error in the cost function without any constraint to avoid trivial  solution. Since we use a good initialization obtained by the CLS algorithm, we do not see any problems in the experiments. Because of the scale problem using displacement error in the full viewing graph loss, which was solve by turning the problem into a PGO in our method, seems unjustified. But using this error in the full viewing graph loss helps getting important insight that becomes clear later in the experiments.


\section{Experiments}\label{sec:Exp.}

To evaluate the proposed algorithms, the challenging real datasets of \cite{wilson2014robust} are used. The camera poses estimations, computed by Bundler \cite{snavely2006photo} and presented in \cite{wilson2014robust} are considered as the ground truth.
In this section, Alg.~\ref{algorithm1}C or Alg.~\ref{algorithm2}C stand for using the displacements error in the proposed algorithms, and Alg.~\ref{algorithm1}O or Alg.~\ref{algorithm2}O stand for using the directions error in each iteration of the algorithms \footnote{Efficient implementation of the proposed algorithms in MATLAB is available via \url{http://visionlab.ut.ac.ir/resources/vgorls.zip}}.

\begin{figure*}
    \includegraphics[width=0.245\textwidth]{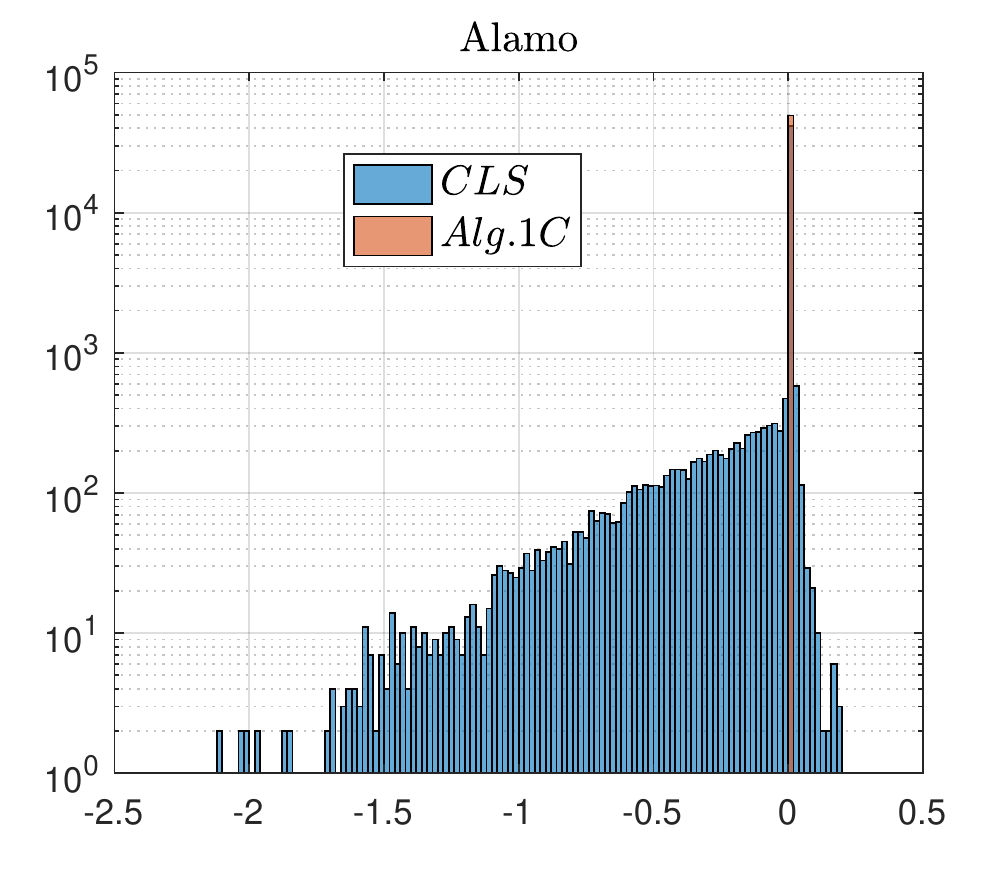}
    \includegraphics[width=0.245\textwidth]{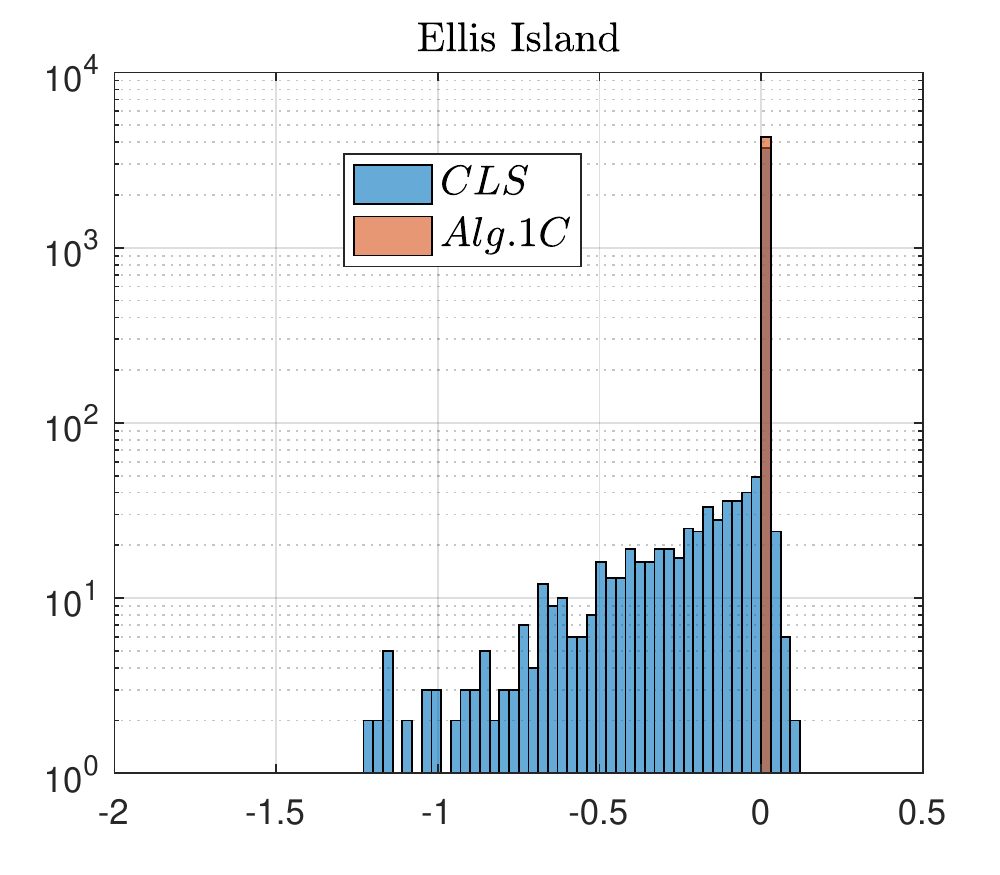}
    \includegraphics[width=0.245\textwidth]{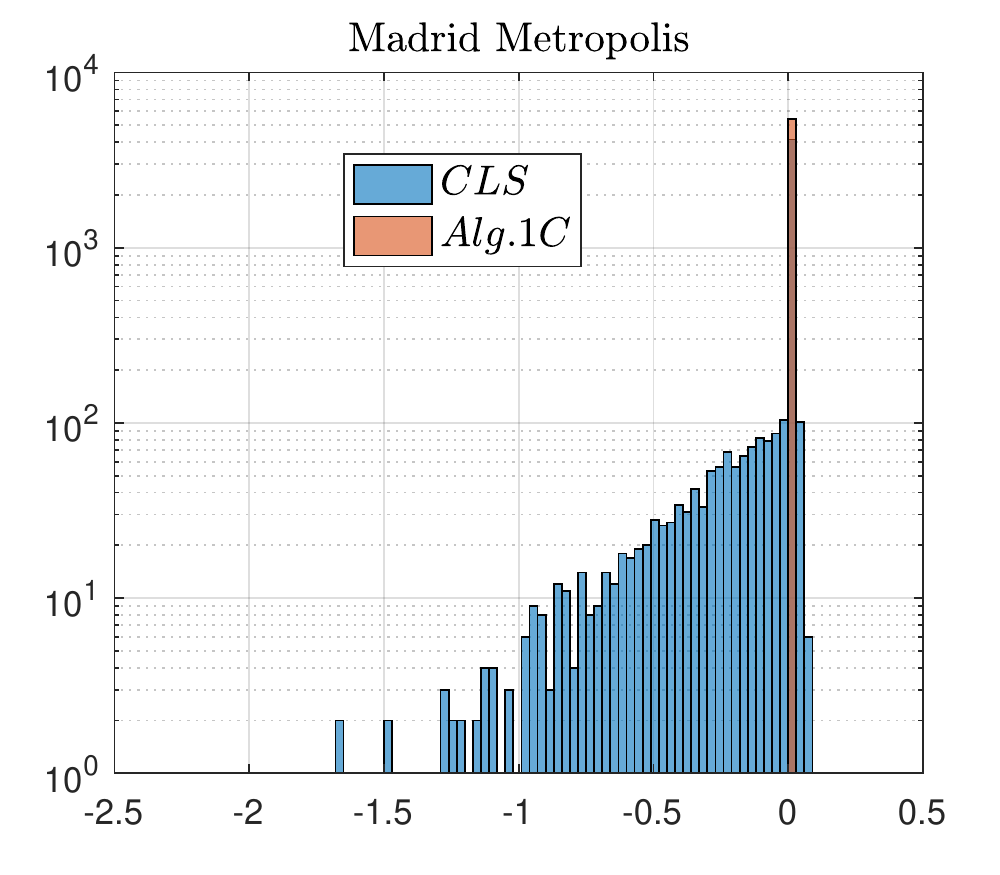}
    \includegraphics[width=0.245\textwidth]{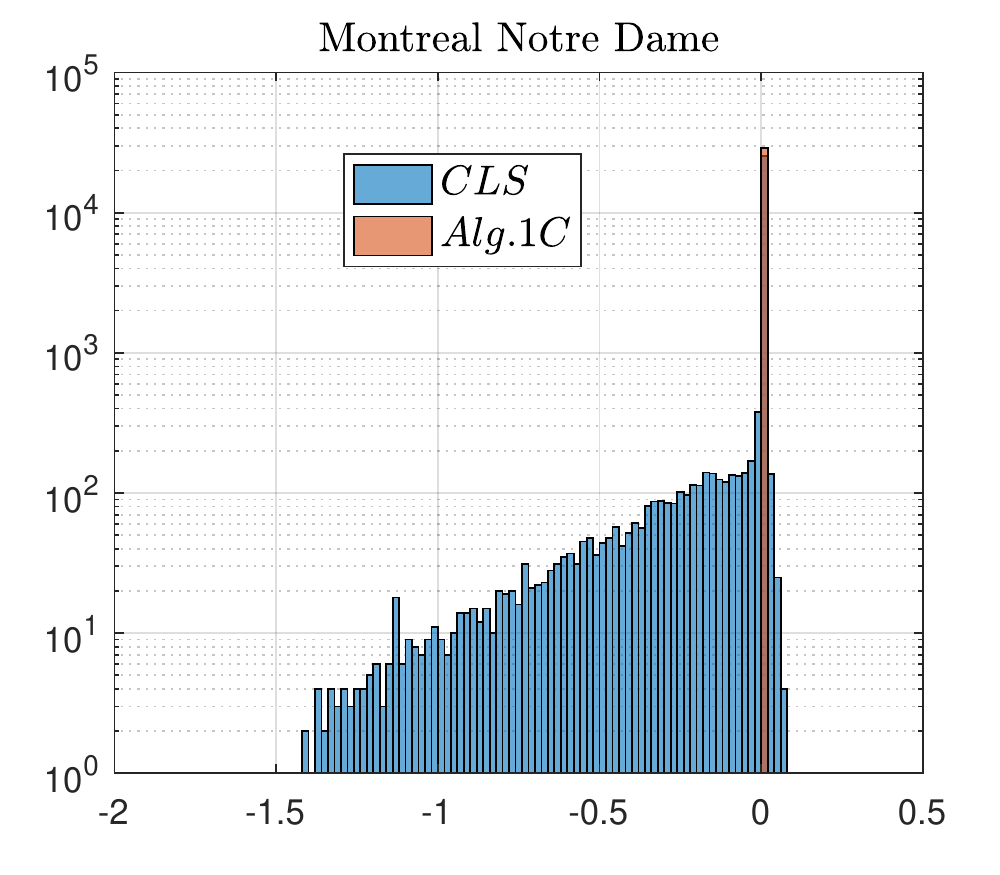}
    \includegraphics[width=0.245\textwidth]{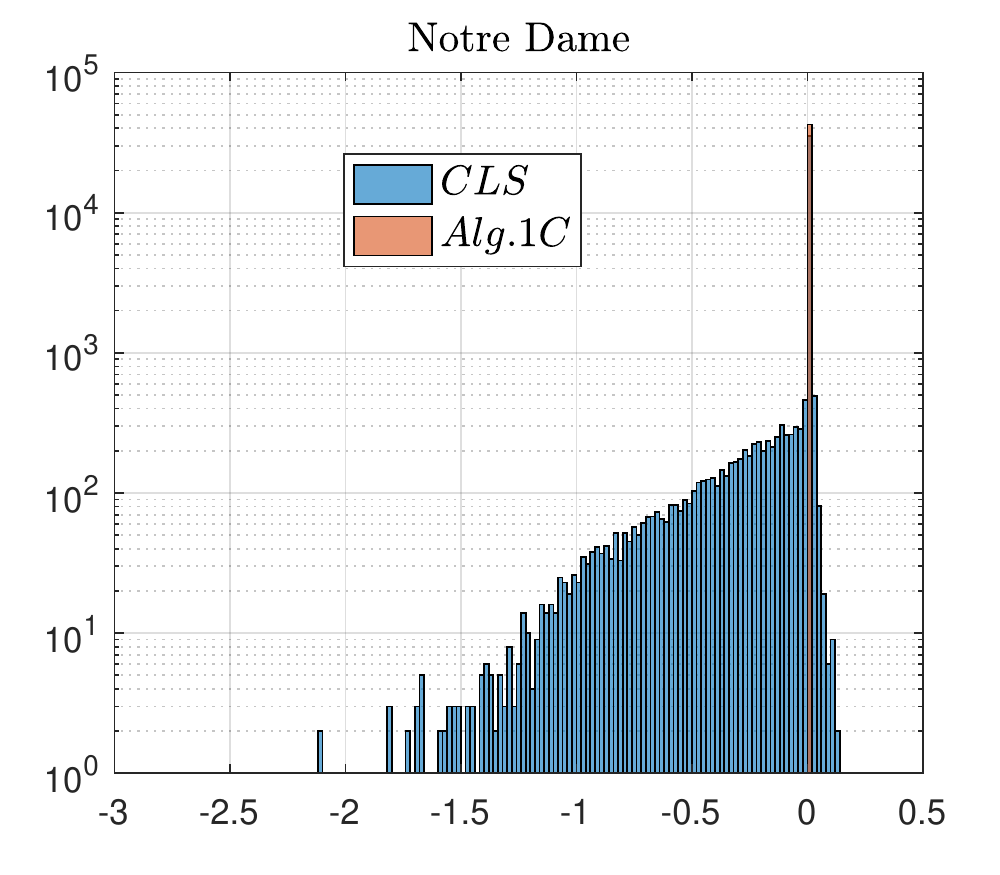}
    \includegraphics[width=0.245\textwidth]{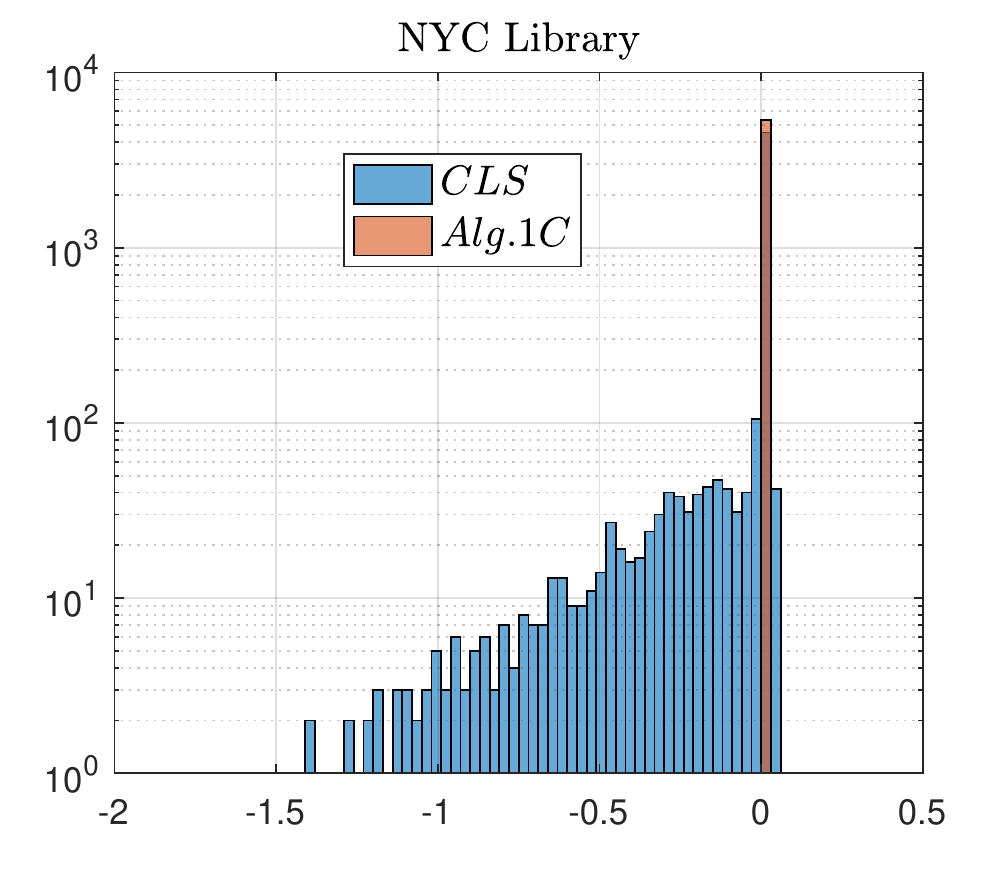}
    \includegraphics[width=0.245\textwidth]{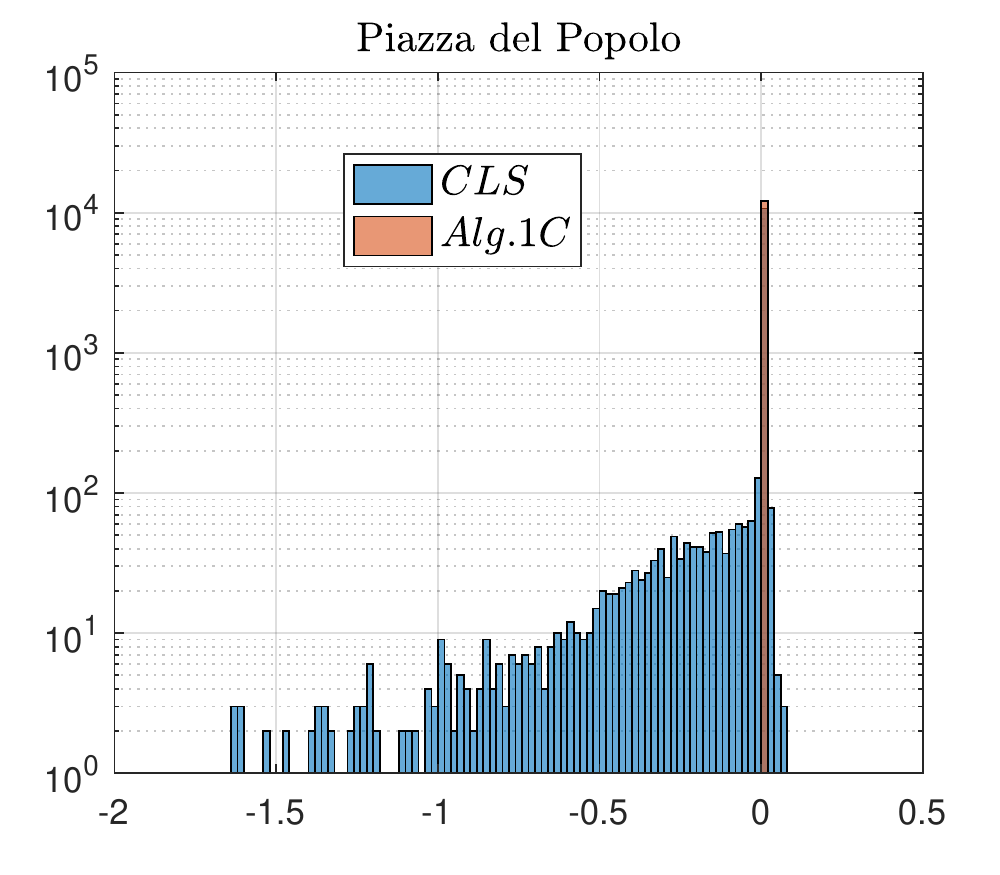}
    \includegraphics[width=0.245\textwidth]{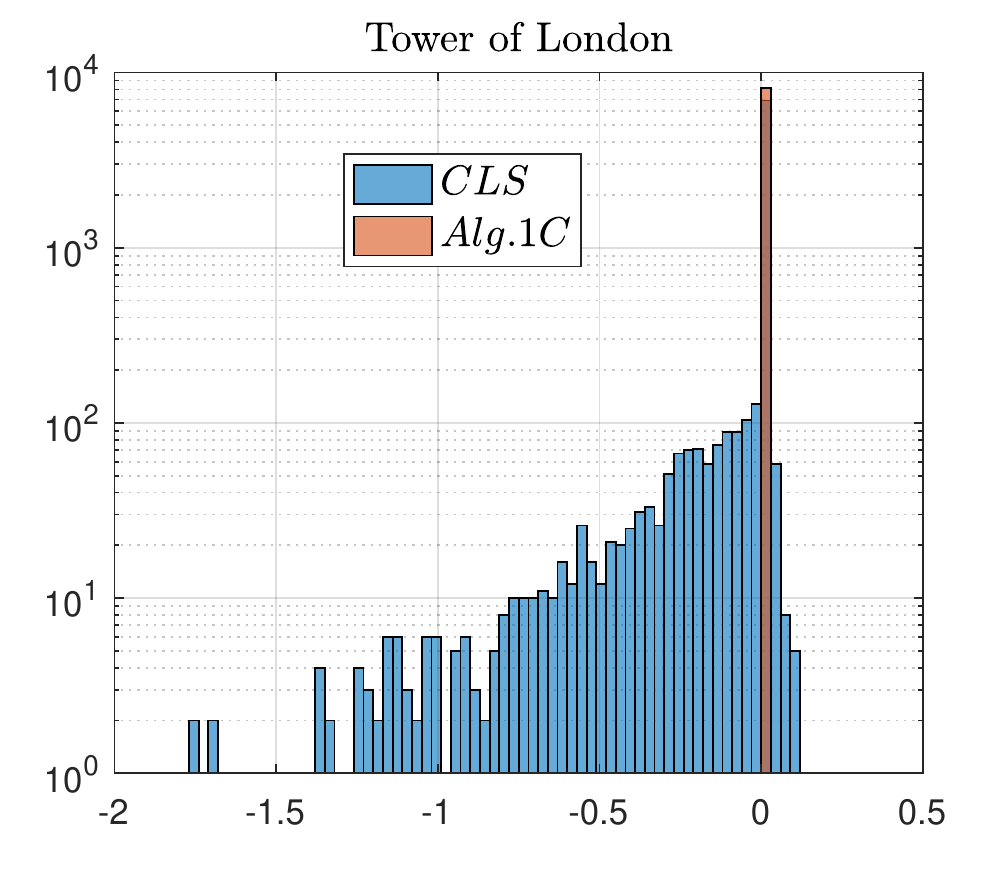}
    \includegraphics[width=0.245\textwidth]{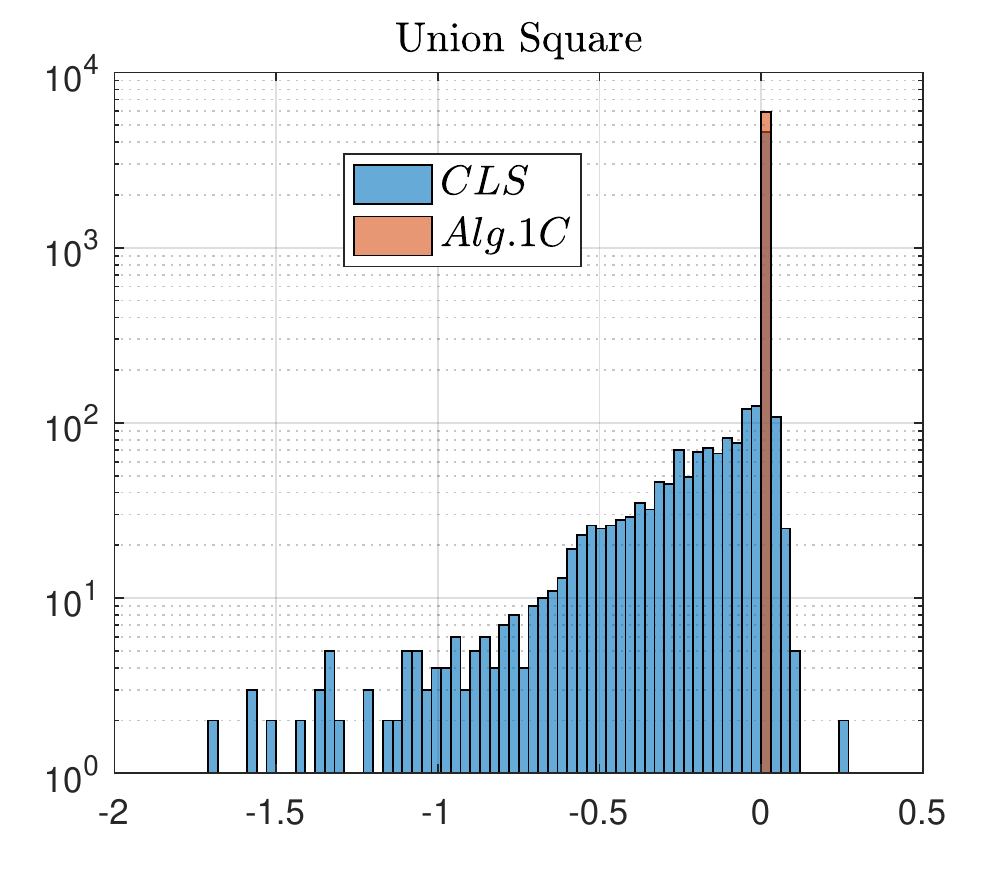}
    \includegraphics[width=0.245\textwidth]{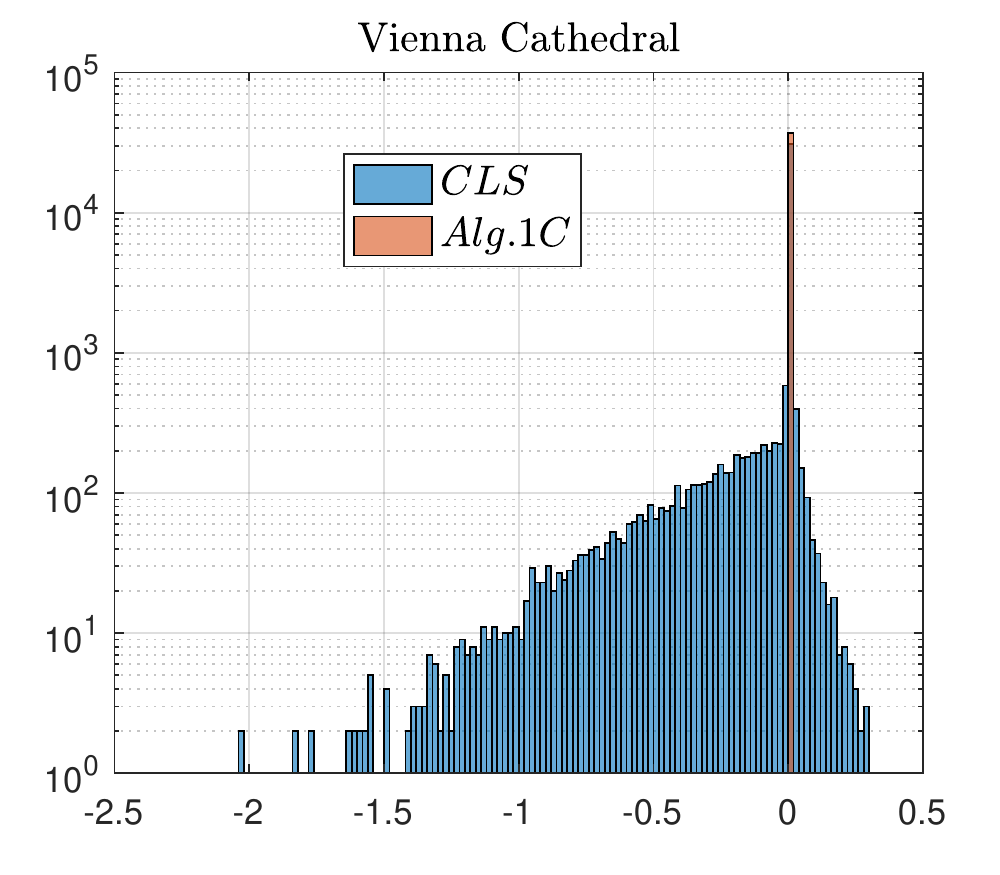}
    \includegraphics[width=0.245\textwidth]{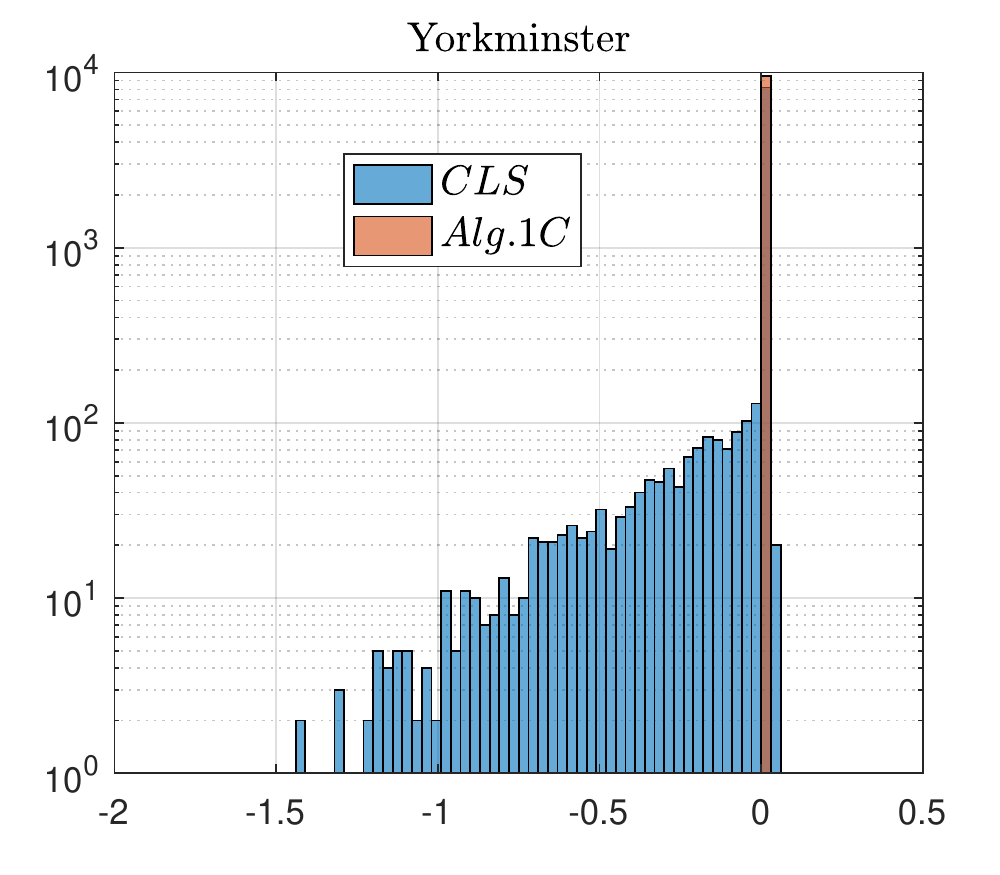}
\caption{Comparing the scale factors obtained by CLS and Alg.~1C on different real datasets. The figures show the histogram of $\log_{10} \frac{\lambda_{ij}}{\|\hat{p}_i-\hat{p}_j\|}$ obtained by the methods on different datasets.}
\label{fig:CLS_Alg1C_all}       
\end{figure*}

For the first step, the weakness of the constraint $\lambda_{ij}\geq1$, which is discussed in section~\ref{conswaek}, is shown experimentally. To this end, the orientation of the cameras are estimated by solving the rotation averaging problem of \eqref{rot_avg} using RS \cite{nasiri2020novel}. Outliers are removed by 1DSFM algorithm, and then the positions are estimated by CLS, i.e. solving \eqref{CLE2} on the constraint of $\lambda_{ij}\geq1$. The results are compared to the results of Alg.~1C.
Fig.~\ref{fig:CLS_Alg1C_all} compares the histogram of the scale factors ratio to the corresponding vertices distances obtained by CLS and Alg.~\ref{algorithm1}C in different datasets. It can be seen that all scale factors obtained by Alg.~\ref{algorithm1}C, despite CLS, match their corresponding vertices distances.

In the following, the recovery performance of the proposed algorithms are evaluated in comparison to the most accurate algorithms proposed in the literature. The comparing methods are Cross method of \cite{govindu2001combining}, CLS method which is used in \cite{tron2009distributed,tron2014distributed}, and LUD method proposed in \cite{ozyesil2015robust}. Since the problem has natural global rotation, translation, and scale ambiguities, an Euclidean transformation should be found that aligns the camera positions to the ground truth as much as possible. Mathematically, we solve
\begin{equation}\label{rts}
    \min_{R,\vt,s} \sum_{i=1}^{n}\left\|\bigl((sR\hat{\vp}_i+\vt)-\vp_i\bigr)\right\|^2,
\end{equation}
where $R$, $\vt$, and $s$ are the rotation matrix, translation vector, and scale parameters of the scaled Euclidean transformation.
The comparison criteria are the mean~($\bar{e}$), the median~($\tilde{e}$), and the root mean squared~($\breve{e}$) distances error. The root mean squared distances error is given by
\begin{equation}\label{RMSE}
    \breve{e} = \sqrt{\frac{1}{m} \sum_{i=1}^m{\|\hat{\vp_i}-\vp_i\|^2}},
\end{equation}
where $m$ is the number of edges.

\begin{figure}
    \includegraphics[width=0.5\textwidth]{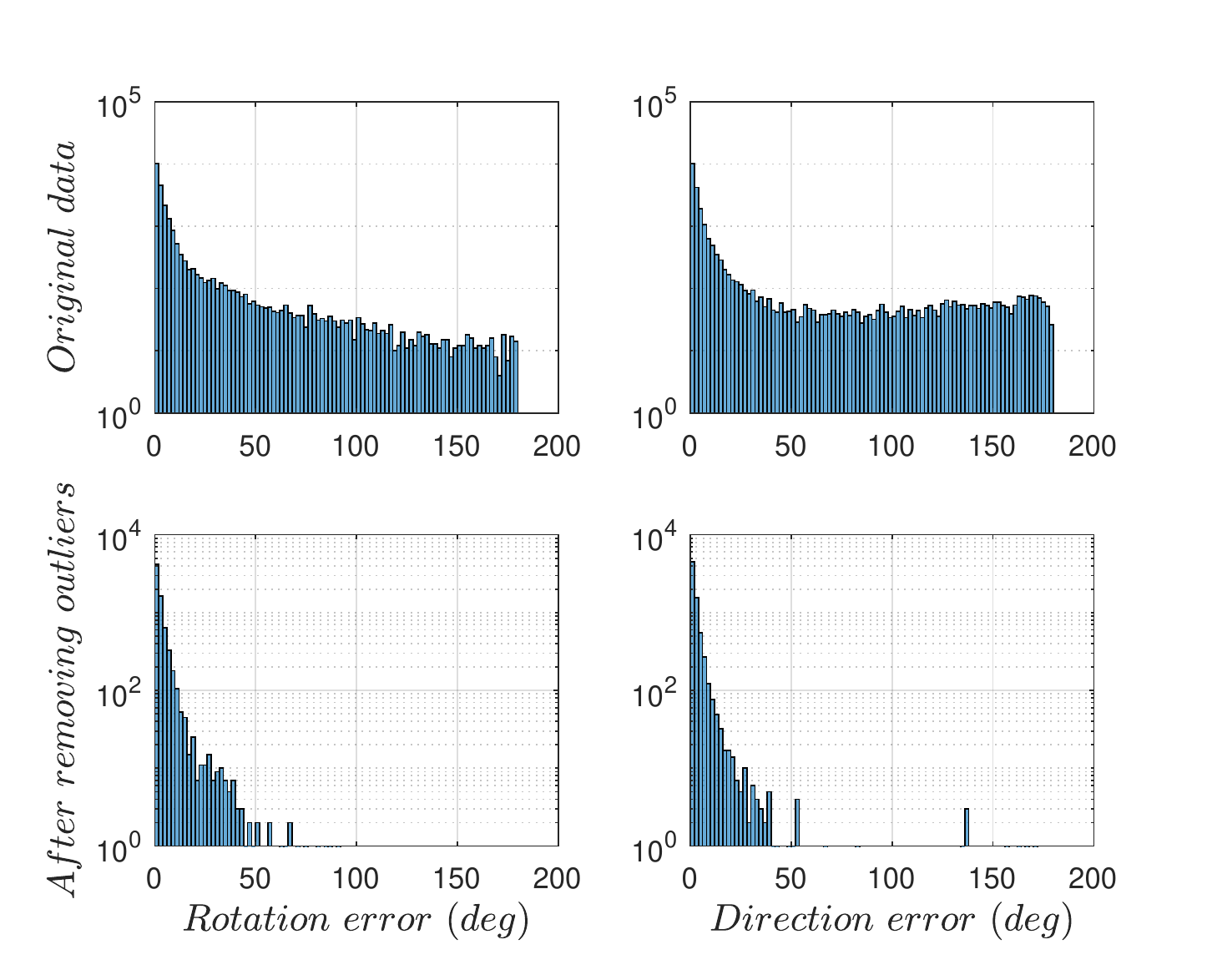}
\caption{Histograms of rotation and direction measurements error, before and after applying the 1DSFM algorithm for ``Tower of London'' dataset.}
\label{fig:hist_Tower_of_London}       
\end{figure}

The histogram of rotation and direction measurements error, before and after applying 1DSFM for \textit{Tower of London} are shown in Fig.~\ref{fig:hist_Tower_of_London}. The figure shows that the most of outliers were removed and the output has a few number of outliers. The similar results were obtained for other datasets.

The results of various methods in different datasets are listed in Table~\ref{tab:1}. The results show that in almost all datasets, the proposed methods outperform the others. Further, the comparison of Alg.~\ref{algorithm1} and Alg.~\ref{algorithm2} demonstrates that minimizing the cost function over rotations and translations simultaneously, results in better camera localization.
Comparing the results of Alg.~\ref{algorithm1}C to Alg.~\ref{algorithm1}O and Alg.~\ref{algorithm2}C to Alg.~\ref{algorithm2}O shows that, although the differences are small, minimizing the displacements error instead of the directions error results in a better camera localization. Because of the scale problem as discussed in the section of the proposed methods, using Alg.~\ref{algorithm2}C is not justified. But interestingly, this algorithm shows the best performance which shows that resulting scales obtained by solving the problem using our proposed method is fine. This shows the importance of using correct weight for combining two terms of the viewing graph optimization problem.


\begin{table*}
\tiny
\setlength{\tabcolsep}{1.4pt}

\caption{Performance comparison of various methods in different original datasets. The table depicts the number of edges, $m$, and the number of vertices, $n$, after removing the outliers using 1DSFM. The comparison criteria are the root mean square error, $\breve{e}$, the mean distance error, $\bar{e}$, and the median distance error ,$\tilde{e}$.}
\label{tab:1}       
\begin{tabular}{|lcc|ccc|ccc|ccc|ccc|ccc|ccc|ccc|}
\hline
\hline\noalign{\smallskip}
\multicolumn{3}{|c|}{Dataset} & \multicolumn{3}{|c|}{Cross} & \multicolumn{3}{|c|}{CLS} & \multicolumn{3}{|c|}{LUD} & \multicolumn{3}{|c|}{Alg.~1C} & \multicolumn{3}{|c|}{Alg.~1O} &
\multicolumn{3}{|c|}{Alg.~2C} & \multicolumn{3}{|c|}{Alg.~2O}\\
Name & m & n & $\breve{e}$ & $\bar{e}$ & $\tilde{e}$ & $\breve{e}$ & $\bar{e}$ & $\tilde{e}$ & $\breve{e}$ & $\bar{e}$ & $\tilde{e}$ & $\breve{e}$ & $\bar{e}$ & $\tilde{e}$ & $\breve{e}$ & $\bar{e}$ & $\tilde{e}$ & $\breve{e}$ & $\bar{e}$ & $\tilde{e}$ & $\breve{e}$ & $\bar{e}$ & $\tilde{e}$\\
\noalign{\smallskip}
\hline\noalign{\smallskip}
                Alamo & 46353 &   479 & 15.94 & 12.96 & 12.52 &  5.36 &  2.89 &  1.78 &  4.89 &  2.56 &  1.43 &  5.46 &  2.57 &  1.18 &  5.39 &  2.61 &  1.30  &  5.14 &  2.17 &  \textbf{0.93} &  \textbf{4.83} &  \textbf{2.11} &  0.95\\
         Ellis Island &  4512 &   193 & 30.85 & 26.18 & 23.26 & 11.19 &  6.87 &  4.22 & 10.92 &  6.17 &  3.57 & \textbf{10.00} &  5.71 &  3.07 & 10.86 &  6.17 &  3.28 & \textbf{10.00} &  \textbf{5.37} &  \textbf{2.65} & 10.29 &  5.50 &  2.68 \\
    Madrid Metropolis &  4150 &   253 & 51.45 & 35.65 & 29.23 & 21.66 & 14.65 & 11.36 & 20.63 & 13.47 & 10.16 & 18.55 & 12.23 &  8.51 & 20.33 & 13.20 & 10.01 & \textbf{16.90} & \textbf{10.95} &  7.05 & 17.28 & 10.98 &  \textbf{6.79}\\
  Montreal Notre Dame & 26330 &   385 & 12.13 & 11.14 & 11.99 &  2.70 &  1.82 &  1.13 &  2.62 &  1.76 &  1.10 &  2.54 &  1.66 &  1.01 &  2.87 &  1.84 &  1.05 &  \textbf{2.45} &  \textbf{1.59} &  \textbf{0.88} &  2.60 &  1.67 &  0.96\\
           Notre Dame & 45048 &   501 & 13.32 &  9.81 &  7.83 &  6.26 &  4.00 &  2.97 &  6.27 &  3.70 &  2.48 &  5.71 &  3.18 &  2.05 &  5.84 &  3.27 &  2.20 & \textbf{5.28} &  \textbf{2.71} &  \textbf{1.60} &  5.39 &  2.78 &  1.68\\
          NYC Library &  5614 &   260 & 15.29 & 13.80 & 14.05 &  4.31 &  3.14 &  2.28 &  \textbf{4.04} &  \textbf{2.82} &  1.88 &  5.06 &  3.55 &  2.41 &  4.60 &  3.15 &  2.05 &  4.85 &  3.03 &  1.92 &  4.87 &  2.96 &  \textbf{1.82}\\
    Piazza del Popolo & 12027 &   234 & 16.43 & 14.29 & 12.56 &  3.30 &  2.24 &  1.48 &  3.10 &  2.03 &  1.30 &  3.04 &  1.95 &  1.17 &  3.31 &  2.11 &  1.18 &  \textbf{2.35} &  \textbf{1.47} &  0.94 &  2.52 &  1.55 &  \textbf{0.91}\\
      Tower of London &  7539 &   371 & 76.41 & 66.35 & 60.20 & 39.01 & 26.75 & 19.02 & 36.52 & 24.32 & 15.37 & 34.16 & 20.49 & 11.86 & 38.13 & 25.15 & 16.79 & \textbf{32.99} & \textbf{18.19} &  \textbf{9.26} & 35.24 & 20.69 & 10.96\\
         Union Square &  5919 &   468 & 18.73 & 14.84 & 10.81 & 12.91 &  9.02 &  6.61 & \textbf{12.67} &  8.68 &  6.27 & 13.17 &  9.07 &  6.74 & 12.89 &  9.02 &  6.89 & 12.80 &  \textbf{8.43} &  \textbf{5.75} & 12.90 &  8.71 &  6.19\\
     Vienna Cathedral & 36729 &   624 & 34.58 & 27.46 & 22.41 & 24.27 & 18.81 & 14.12 & 22.84 & 17.62 & 13.17 & 20.90 & 15.59 & 10.85 & 25.45 & 19.20 & 13.15 & \textbf{19.56} & \textbf{14.42} & \textbf{10.06} & 23.32 & 17.29 & 11.99\\
          Yorkminster & 10249 &   317 & 27.97 & 13.74 &  5.94 & 15.99 &  8.68 &  5.77 & 17.09 &  7.85 &  4.99 & 15.00 &  7.64 &  4.74 & 14.89 &  7.45 &  4.81 & 14.39 &  6.88 &  4.05 & \textbf{14.27} &  \textbf{6.70} &  \textbf{3.81}\\
\noalign{\smallskip}\hline
\hline
\end{tabular}
\end{table*}

\section{Conclusion}\label{sec:Conclusion}
In this paper,  we proposed iterative methods for solving the camera location estimation problem. One of our proposed methods fixed the unknown scale factors, i.e. $\lambda_{ij}$s, which appear on the common cost function \eqref{CLE2}, in each iteration. Hence, the solution is obtained through solving iterative least-squares problems. The scale factors in each iteration are set to the distance between the corresponding vertices distances in the previous iteration. We observed experimentally that this helps the final scale factors to converge to the distances of the corresponding vertices, i.e. $\lambda_{ij}\rightarrow\|\hat{\vp}_j-\hat{\vp}_i\|$. This solved the problem of CLS algorithm that bunches up a significant number of scale factors in the lower bound.

Fixing the parameters $\lambda_{ij}$s converts the direction observations to translation observations and converts the given viewing graph to a Pose-Graph. This motivated us to use the state-of-the-art method of PS \cite{nasiri2020novel} to obtain rotations and positions of cameras simultaneously and to propose the second algorithm. Obtaining the best results by solving the full cost of viewing graph optimization problem is an important result. It is an step toward to solve challenging viewing graph optimization problem and calls other researches to put emphasis on developing the algorithms considering both terms together in the viewing graph optimization problem. 

Both algorithms can solve the original cost function in which the chordal distance of directions are minimized. But the experiments show that minimizing the common used displacement distance metric (weighting the directions error by the edges length), results in a better performance in real datasets. This is an interesting result, because using displacement distance metric has scale problem which can be more important in the case of the full viewing graph optimization solved using our Alg. 2C. Having the best performance of Alg. 2C shows the importance of having good weights for combining the two term of the viewing graph optimization problem. An interesting future direction of work is obtaining the covariance of noise for using accurate weights for combining two terms in the loss function of viewing graph optimization.

We observed experimentally that all our proposed algorithms converge. Another important line of future research that we are undertaking is developing theoretical convergence results for our proposed algorithms.

\section*{Appendix}
The author in \cite{govindu2001combining} proposed an iterative reweighted algorithm for solving orthogonal distance of directions error, i.e. the solution of \eqref{CLE1} with orthogonal distance.  
%
%
The strategy is to minimize orthogonal distance of displacements error which can be solved in the closed-form and then reweigthing errors by the squared distance of the edges. Although upon convergence, the algorithm yields that the weights equal to the edges lengths, there is no guarantee that the algorithm converges.

We use a novel formulation in which the solution is directly the minimizer of \eqref{CLE1} where the distance metric is orthogonal distance. We use the cost function of~\eqref{Crs3} that uses $\lambda_{ij}$ multipliers before the second term. The problem \eqref{Crs3} does not need any constraint and the minimizer of the cost function is directly the solution of \eqref{CLE2} with the orthogonal distance metric.

Suppose that the set of $\{\lambda_{ij}^*,p_i^*\}$ is the optimal solution of \eqref{Crs3}. So we have,
\begin{multline}\label{neq}
\forall e_{ij}=\{i,j\}\in E, \forall \lambda_{ij}\in \mathbb{R},\\
    \left\|\gamma_{ij}-\lambda_{ij}^*(\vp_j^*-\vp_i^*)\right\|^2 <
    \left\|\gamma_{ij}-\lambda_{ij}(\vp_j^*-\vp_i^*)\right\|^2.
\end{multline}

Since $\lambda_{ij}(\vp_j^*-\vp_i^*)$ comprises all points on the line which connects $\vp_i$ to $\vp_j$, it can be conclude from \eqref{neq} that $\lambda_{ij}^*(\vp_j^*-\vp_i^*)$ is the nearest point of $\vp_i\vp_j$-line to the endpoint of $\gamma_{ij}$. Obviously the nearest point obtained by projecting $\gamma_{ij}$ onto $\vp_i\vp_j$-line, and its Euclidean distance is equal to cross product of $\gamma_{ij}$ and $\frac{\vp_j^*-\vp_i^*}{\|\vp_j^*-\vp_i^*\|}$.  

Problem \eqref{Crs3} can be solved by a two block coordinate descent approach. The algorithm repeats the following two steps. In one step, the cost is optimized over $\vp_i$s while $\lambda_{ij}$s are fixed, and in the next step the cost is optimized over $\lambda_{ij}$s while $\vp_i$s are fixed vectors. Both steps are linear least-squares problem and can be solved in a closed form.
It is easy to show that the proposed coordinate descent algorithm satisfies the convergence conditions of theorem~1 of \cite{rouzban2019rate}. Therefore, the algorithm finds the minimizer of the sum of squared orthogonal distance of directions error. Although we were able to prove a convergence result for minimizing orthogonal distance, we obtained bad results using this method in experiments.


\bibliographystyle{spmpsci}
\bibliography{ref}

\end{document}